# Instruction Multi-Constraint Molecular - Generation Using a Teacher-Student Large Language Model

Peng Zhou[1], Jianmin Wang[2], Chunyan Li[3], Zixu Wang[4], Yiping Liu[1], Siqi Sun[5,6], Jianxin Lin[1],

Leyi Wei[7,8], Xibao Cai[1], Houtim Lai[9], Wei Liu[9], Longyue Wang[9,*], Yuansheng Liu[1,*],

Xiangxiang Zeng[1,*]

[1]College of Information Science and Engineering, Hunan University, Changsha, China

[2] The Interdisciplinary Graduate Program in Integrative Biotechnology, Yonsei University, Incheon 21983, Korea

[3] School of Informatics, Yunnan Normal University, Kunming 650500, China

[4] Department of Computer Science, University of Tsukuba, Tsukuba 3058577, Japan

[5]Research Institute of Intelligent Complex Systems, Fudan University, Shanghai, 200433, China.

[6]Shanghai AI Laboratory, Shanghai, 200232, China.

[7]Centre for Artificial Intelligence driven Drug Discovery, Faculty of Applied Science, Macao Polytechnic University, Macao SAR, China

[8]School of Informatics, Xiamen University, Xiamen, China

[9]Tencent AI Lab

*Corresponding authors: vinnylywang@tencent.com, {yuanshengliu, xzeng}@hnu.edu.cn

# Abstract

While various models and computational tools have been proposed for structure and property analysis of molecules, generating molecules that conform to all desired structures and properties remains a challenge. Here, we introduce a multi-constraint molecular generation large language model, TSMMG, which, akin to a student, incorporates knowledge from various small models and tools, namely, the 'teachers'. To train TSMMG, we construct a large set of text-molecule pairs by extracting molecular knowledge from these 'teachers', enabling it to generate novel molecules that conform to the descriptions through various text prompts. We experimentally show that TSMMG remarkably performs in generating molecules meeting complex, natural language-described property requirements across two-, three-, and four-constraint tasks, with an average molecular validity of over 99% and success ratio of 82.58%, 68.03%, and 67.48%, respectively. The model also exhibits adaptability through zero-shot testing, creating molecules that satisfy combinations of properties that have not been encountered. It can comprehend text inputs with various language styles, extending beyond the confines of outlined prompts, as confirmed through empirical validation. Additionally, the knowledge distillation feature of TSMMG contributes to the continuous enhancement of small models, while the innovative approach to dataset construction effectively addresses the issues of data scarcity and quality, which positions TSMMG as a promising tool in the domains of drug discovery and materials science.

# Background

The development and application of molecular generation models play an essential role in the field of Artificial Intelligence for Drug Discovery (AIDD). Molecular generation models are instrumental in addressing the challenges and complexities associated with the identification and design of novel therapeutic compounds. In contrast to traditional virtual screening approaches, involving the sift of desired molecules from existing libraries, these innovative models are engineered to directly generate novel molecules. Their ability to navigate vast chemical spaces, optimize lead compounds, and facilitate de novo design positions them as indispensable tools in the pursuit of novel and effective therapeutic interventions [1][2][3][4][5]. These models not only exhibit the ability to generate chemically valid molecules that precisely adhere to the requirements of molecular analysis tools [6][7], but they also excel in the generation of molecules that meet specific constraints, like quantitative estimate of drug-likeness (QED) and molecular hydrophobicity (LogP) [8][9].

However, a primary challenge in the realm of drug discovery lies in identifying molecules that conform to a multitude of constraints, including binding affinity, LogP, QED, synthetic accessibility (SA), and toxicity, among others, rather than merely generating compounds that are chemically valid or solely meeting specific criteria [10][11]. Several works have been introduced to address this challenge, presenting methodologies capable of generating molecules that adhere to a spectrum of concurrent condition constraints. For instance, Li et al. introduced a conditional generative model proficient in generating molecules that meet both SA and QED criteria, even yielding dual-target inhibitors for JNK3 and GSK3 [12]. Jin et al. achieved this feat by extracting diverse substructures with varying properties and reassembling them to produce molecules satisfying QED, SA, and the inhibition of both JNK3 and GSK3 [13]. Bagal et al. employed a transformer decoder architecture, treating constraint conditions as conditional codes, to explore the generation of molecules under various combinations of multiple constraints, including LogP, TPSA (Total Polar Surface Area), and SA [14]. Wang et al. utilized a combination of a conditional transformer, knowledge distillation, and reinforcement learning to generate molecules with activity against DRD2, while also ensuring adherence to QED and SA criteria [10].

Although significant progress has been made in prior endeavors, it is important to acknowledge that multi-constraint molecular generation methods still suffer from several noteworthy limitations, which hinder their practical applicability. These limitations undermine the overall effectiveness and efficiency of these methods in generating molecules that simultaneously meet diverse sets of constraints in drug discovery. These limitations fall into the following points: (1) **Current multi-constraint molecular generation methods heavily rely on a narrow set of constraints.** These methods predominantly focus on specific molecular properties, such as LogP, QED, SA, DRD2, JNK3, and GSK3. As a result, they may overlook other crucial aspects like substructures, bioavailability, and toxicity. The restricted range of constraints limits the comprehensive exploration of diverse chemical properties, potentially hindering the discovery and optimization of molecules with broader applicability in drug discovery and related domains. (2) **These methods often require extensive fine-tuning when applied to different tasks.** They tend to generate molecules that closely adhere to the feature distribution of the training dataset. As a consequence, adapting these models to changes in the target space or applying them to diverse tasks necessitates significant retraining. This inflexibility makes the models less adaptable, introducing a substantial burden in terms of computational resources and time when confronted with variations in the application context. (3) **They often involve intricate designs.** The complexity of the models and algorithms used can be a significant obstacle in their practical application. Users may find it challenging to understand and navigate the complexities of the methods, impacting their usability. Improving the simplicity of these models is essential to make them more accessible and applicable in real-world scenarios, especially in drug discovery and related domains.

To address the challenges, we introduce the **T**eacher-**S**tudent-based **M**ulti-constraint **M**olecular **G**eneration (TSMMG) model, a natural language-based multi-constraint molecular generation approach. TSMMG offers several pivotal advantages: (1) **Flexible Data Generation Framework**: Our approach presents a versatile data generation framework that leverages a range of molecular tools and advanced models to selectively extract molecules with diverse properties from publicly available molecular libraries. The framework is based on the concept of knowledge distillation, where various tools and models used to extract molecular knowledge are referred to as 'teachers', while our model, TSMMG, is referred to as the 'student'. Any molecular-related tool or model can serve as a teacher, providing diverse molecular knowledge to TSMMG. The utilization of the teacher-student paradigm in our model provides a highly scalable approach, facilitating the seamless absorption of molecular knowledge beyond the scope of this paper. Moreover, this approach can be easily extended to other domains, such as materials science. (2) **Multi-task capability**: Harnessing the capabilities of large language models, we can concurrently train across multiple tasks. By formulating distinct prompts, we delineate unique molecular spaces without the need for repetitive fine-tuning. This strategy capitalizes on the adaptability of large language models. (3) **Simple Architecture**: The proposed model adopts a transformer-based decoder architecture. This design, characterized by its simplicity, eliminates the need for intricate preprocessing of molecular data.

To showcase the expressive capabilities of our proposed model, we meticulously designed 16 sets of experiments for multi-constraint molecular generation. These experiments covered a spectrum of tasks, including molecular substructures, physicochemical properties, affinity with targets, and ADMET properties. Our findings from these experiments are compelling: TSMMG not only yields over 99% of legally valid molecules based on natural language instructions but also, notably, a substantial proportion of these molecules impeccably aligns with the specified properties in the textual descriptions. Furthermore, we conducted a noteworthy case study involving a zero-shot 5-constraint task. In this scenario, TSMMG successfully produced molecules capable of simultaneous binding to EP2 and EP4, showcasing favorable drug-likeness and synthetic accessibility, along with the ability to penetrate the blood-brain barrier. This case study serves as an additional testament to the vast potential embedded in TSMMG. Additionally, our model demonstrated its prowess in understanding natural language beyond the prompts outlined in this paper, as empirically validated. This expanded capability further solidifies the model's practical applicability. Moreover, we observed that integrating novel molecules generated by our model significantly enhances the teacher model's performance. This collaborative synergy fosters continuous improvement between the teacher and student models, underscoring the model's adaptability and potential for iterative refinement.

# Result

## TSMMG approach

As shown in Figure 1, TSMMG process involves the following steps: (1) To begin, a substantial dataset of molecules is collected from publicly available molecular libraries. This dataset undergoes analysis by advanced molecular parsing tools and models, which referred to as "teachers". These teachers extract extensive information, encompassing structural details, physicochemical properties, binding affinities to various targets, and other pertinent attributes for each molecule. The resulting knowledge is then organized into natural language descriptions, which are paired with the corresponding molecules. (2) Next, the "student" model, TSMMG, is introduced and trained using the knowledge obtained in the previous step. TSMMG is designed to create a direct mapping from natural language descriptions to molecular language. By absorbing a diverse range of knowledge expressed in natural language, the model acquires the capability to generate molecules that possess the specified properties outlined in the text. It's worth noting that TSMMG undergoes pre-training on a vast corpus of pure text, enabling it to effectively understand and interpret natural language. (3) When presented with a text description that includes multiple constraints, TSMMG can generate entirely novel molecules that fulfill these textual descriptions. In doing so, it effectively bridges the gap between natural language and molecular language for the purpose of multi-constraint molecular generation.

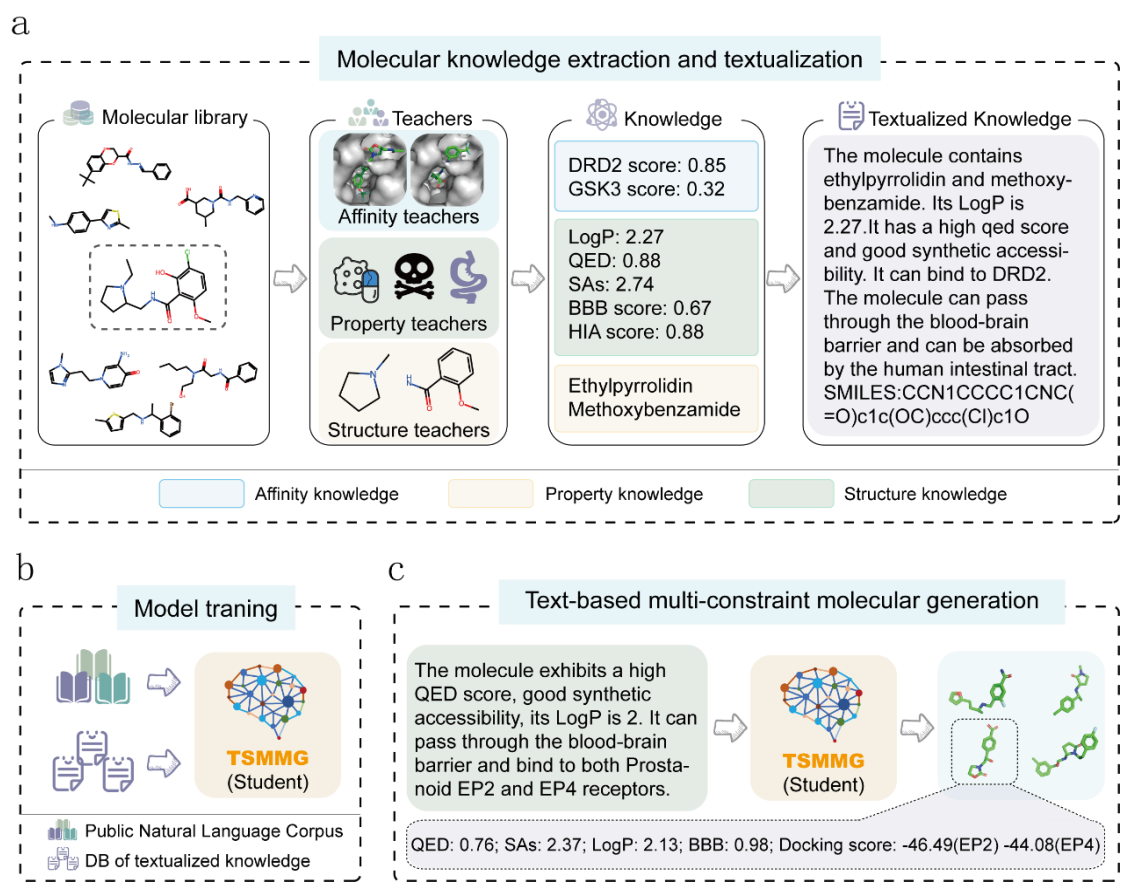

**Figure 1** The process of TSMMG is illustrated as follows: (a) In the initial step, a comprehensive dataset of molecules is gathered from publicly available molecular libraries. This dataset is then subjected to advanced molecular parsing tools and models, which referred to as "teachers." These teachers extract a wealth of information, including structural details, physicochemical properties, binding affinities to various targets, and other relevant attributes for each molecule. The acquired knowledge is then structured into natural language descriptions, resulting in a substantial dataset containing paired natural language descriptions and their corresponding molecules. (b) The "Student" model, TSMMG, is introduced and trained using the knowledge acquired in the previous step. TSMMG learns to directly map from natural language descriptions to molecular language. By absorbing diverse knowledge expressed in natural language, the model gains the ability to generate molecules that possess the specified properties detailed in the text. Note that TSMMG undergoes pre-training on a vast corpus of pure text, which enables it to comprehend and interpret natural language effectively. (c) When presented with a text description containing multiple constraints, TSMMG can generate entirely new molecules that satisfy these textual descriptions, effectively bridging the gap between natural language and molecular language for multi-constraint molecular generation.

# Multi-constraint tasks

## Task setting

To comprehensively demonstrate the efficacy of the TSMMG model, we categorized multi-constraint tasks into three types, each based on different levels of complexity: two-constraint molecular generation, three-constraint molecular generation, and four-constraint molecular generation. Each of these three task categories comprises eight one-constraint tasks. These one-constraint tasks encompass:

**Structure Task:**

Task 1. Specifying a functional group (FG).

**Physicochemical Property Tasks:**

Task 2. Specifying the level of hydrophilicity and hydrophobicity (LogP = 1).

Task 3. Specifying the level of quantitative estimate of drug-likeness (QED > 0.6).

Task 4. Specifying the level of synthetic accessibility score (SAs < 4).

**Activity Tasks:**

Task 5. Generating molecules with high affinity for the dopamine type 2 receptor (DRD2 > 0.5).

Task 6. Generating molecules with high affinity for the glycogen synthase kinase-3 beta (GSK3 > 0.5).

**ADMET Property Tasks:**

Task 7. Generating molecules capable of crossing the blood-brain barrier (BBB > 0.5).

Task 8. Generating molecules that can be absorbed by the human small intestine (HIA > 0.5).

We employ the '+' symbol to concatenate multiple one-constraint tasks, thereby representing multi-constraint tasks. Within the two-constraint tasks, we considered eight subtasks, combining structure tasks with individual physicochemical property tasks, activity tasks, and ADMET property tasks. These include (1) FG+FG, 2FG for short; (2) FG+LogP; (3) FG+QED; (4) FG+SAs; (5) FG+DRD2; (6) FG+GSK3; (7) FG+BBB; and (8) FG+HIA. In the three-constraint tasks, we explored subtasks such as (1) FG+DRD2+QED; (2) FG+GSK3+QED; (3) FG+BBB+QED; and (4) FG+HIA+QED. The four-constraint tasks include (1) FG+DRD2+QED+SAs; (2) FG+GSK3+QED+SAs; (3) FG+BBB+QED+SAs; and (4) FG+HIA+QED+SAs.

It is essential to emphasize that all these tasks were completed within a single model, employing different natural language prompts. The model underwent comprehensive training in a unified process, eliminating the need for repetitive fine-tuning. The specific prompts used in these experiments are detailed in Table 1.

**Table 1** The prompts we use in this work. [FG], [FG1] and [FG2] refer to any functional group, and [VALUE] refers to a real number.

| task | prompt |
|---|---|
| 2FG | The molecule contains [FG1],[FG2]. |
| FG+LogP | The molecule contains [FG]. Its LogP is [VALUE]. |
| FG+QED | The molecule contains [FG]. It has a high qed score. |
| FG+SA | The molecule contains [FG]. It has good synthetic accessibility. |
| FG+DRD2 | The molecule contains [FG]. It is active to DRD2. |
| FG+GSK3 | The molecule contains [FG]. It is active to GSK3. |
| FG+BBB | The molecule contains [FG]. It can pass through the blood-brain barrier. |
| FG+HIA | The molecule contains [FG]. It can be absorbed by human intestinal. |
| FG+DRD2+QED | The molecule contains [FG]. It is active to DRD2. It has a high qed score. |
| FG+GSK3+QED | The molecule contains [FG]. It is active to GSK3. It has a high qed score. |
| FG+BBB+QED | The molecule contains [FG]. It can pass through the blood-brain barrier. It has a high qed score. |
| FG+HIA+QED | The molecule contains [FG]. It can be absorbed by human intestinal. It has a high qed score. |
| FG+DRD2+QED+SAs | The molecule contains [FG]. It is active to DRD2. It has a high qed score. It has good synthetic accessibility. |
| FG+GSK3+QED+SAs | The molecule contains [FG]. It is active to GSK3. It has a high qed score. It has good synthetic accessibility. |
| FG+BBB+QED+SAs | The molecule contains [FG]. It can pass through the blood-brain barrier. It has a high qed score. It has good synthetic accessibility. |
| FG+HIA+QED+SAs | The molecule contains [FG]. It can be absorbed by human intestinal. It has a high qed score. It has good synthetic accessibility. |
| BTK | The molecule can bind to BTK. |
| FGFR4 | The molecule can bind to FGFR4 |
| KPCD3 | The molecule can bind to KPCD3. |
| 3CL | The molecule can bind to 3CL. |

# Performance Analysis

The experimental results, depicted in Figure 2-a and Figure 2-b, unveil several noteworthy findings: **Validity**: The model demonstrates a remarkable ability to generate molecules that adhere to the syntax rules of SMILES (Simplified Molecular Input Line Entry System) [15], with an impressive average validity rate of 99.87%, 99.89%, and 99.87% for two constraint tasks, three constraint tasks, and four constraint tasks, respectively. This underscores the model's proficiency in consistently producing grammatically correct molecules. **Uniqueness**: Most generated molecules are unique, with an outstanding average uniqueness rate of 90.27%, 81.2%, and 81.89% for two constraint tasks, three constraint tasks, and four constraint tasks, respectively. From a specific task perspective, the uniqueness of tasks related to DRD2 and GSK3 is relatively low, averaging less than 70%, while other tasks score above 90%. In the next section, we will analyze the reasons behind this situation. Overall, the model consistently generates largely distinct molecules across various tasks. **Novelty**: The average novelty of the generated molecules stands at 92.79%, 87.6%, and 87.87%. Similar to uniqueness, tasks related to DRD2 and GSK3, such as FG+DRD2+QED (82.76%), FG+GSK3+QED (82.44%), FG+DRD2+QED+SA (83.9%), and FG+GSK3+QED+SA (83.14%), have relatively lower novelty scores, while other tasks have novelty scores exceeding 90%. In general, the model demonstrates a capacity to generate innovative molecules for most of the tasks at hand. **Diversity:** Most generated molecules exhibit notable structural differences, as reflected in the outstanding average diversity score of 90.47, 89.3, and 89.37. Similarly, tasks related to DRD2 and GSK3 exhibit lower diversity compared to other tasks. **Success Rates**: The average success rate stands at 82.58%, 68.03%, and 67.48% for two constraint tasks, three constraint tasks, and four constraint tasks, respectively, which demonstrates the model's efficacy in generating novel molecules that effectively meet all requirements specified by natural language.

## Impact of FGs on Performance

An important distinction from previous methods is that we consider functional groups (FG) as an additional constraint, allowing for more precise control over the direction of generation. Given the relatively low uniqueness in tasks related to DRD2 and GSK3, we use the FG+DRD2 task as an example to further discuss the impact of FG on generation results.

Firstly, we analyze the reasons for the low uniqueness (68.48) of the FG+DRD2 task. In this task, our prompt template is "The molecule contains [FG]. It is active to DRD2." We randomly selected 1000 FGs to form 1000 prompts, each generating 5 molecules, totaling 5000 molecules. The only difference between each prompt is the FG, so we group according to the number of unique molecules generated by each prompt and then extract the FG from these prompts for analysis. As shown in Figure 2-c-(1), we see that the number of FGs that led to the generation of 1, 2, 3, 4, and 5 unique molecules were 195, 287, 252, 184, and 82, respectively. Over 80% of the FG-associated prompts can generate two or more unique molecules, with 82 FG-associated prompts each generating 5 completely different molecules. A significant portion of FG-associated prompts tend to generate identical molecules. We speculate that the main reason for this is the inconsistent frequency of these FGs in the training set, causing the model to be unable to effectively learn the larger molecular space corresponding to the FG. In light of this speculation, we grouped these 1000 FGs according to the number of unique molecules generated and calculated the average frequency of the FGs in the training set within each group. As shown in Figure 2-c-(2), this is basically consistent with our speculation. Except for the group generating one unique number of molecules as group 1, as the number of unique molecules generated increases, the frequency of the corresponding group's FG in the training set also increases, indicating that these FGs can be better trained. The average frequency of the FGs in the group with one unique number is slightly higher than that of the group with a 2 unique number, and lower than the other groups, which we assume is an acceptable bias. We then checked the frequency of the FGs in the DRD2 related training set. As shown in Figure 2-c-(3), a considerable portion of functional groups (FGs) did not appear in the training set related to DRD2. Despite this, our model still demonstrates the capability to generate correct molecules.

Further, we consider the ratio of molecules that simultaneously satisfy Valid, Unique, Novelty, and Success (VUNS) criteria generated by different groups. As shown in Figure 2-c-(4), combined with Figure 2-c-(2) and Figure 2-c-(3), as the frequency of FGs in the training set increases, the model is more capable of generating more novel molecules that meet the constraints. In group 5, the average frequency of this group's FG in all training set is 1339, and the ratio of VUNS molecules generated by these FG-associated prompts is as high as 91%.

Given the significant impact of FG on the success rate, we calculated the success ratio without considering FG matching, abbreviated as SR (nFG). For example, for the FG+DRD2+QED+SA task, SR (nFG) only considers whether DRD2, QED, and SA meet the constraints. The results are shown in Figure 2-d. It can be seen that in the two-constraint task, three-constraint task, and four-constraint task, the success rate without considering FG is 13.06%, 16.51%, and 16.76% higher than the success rate considering FG, respectively.

The above observations suggest that as more molecules and FGs are added to the training set, our model should be able to achieve more significant performance.

## Effect of Temperature on Performance

During the inference process of large language models, temperature is a parameter of great interest. A lower temperature implies lower randomness, while a higher temperature means the model has greater freedom. We conducted tests on all tasks by setting different temperatures. Figure 2-e shows the average performance of all tasks when the temperature is set to 0.5, 0.75, 1.0, 1.25, and 1.5, respectively. It can be observed that as the temperature increases, the ability to generate valid molecules remains virtually unchanged, still maintaining above 99%. Novelty and diversity also remain almost unchanged. However, unique and SR show a noticeable increase or decrease. Unique increases from 73.42% to 90.04%, an improvement of approximately 17%, indicating that as the temperature increases, the model can generate more unique molecules. At the same time, SR decreases from 83.03% to 61.94%, a reduction of about 21%. The decrease in SR is roughly consistent with the increase in unique, which means that although increasing the temperature from 0.5 to 1.5 generates 17% more unique molecules, most of them do not satisfy all constraint conditions.

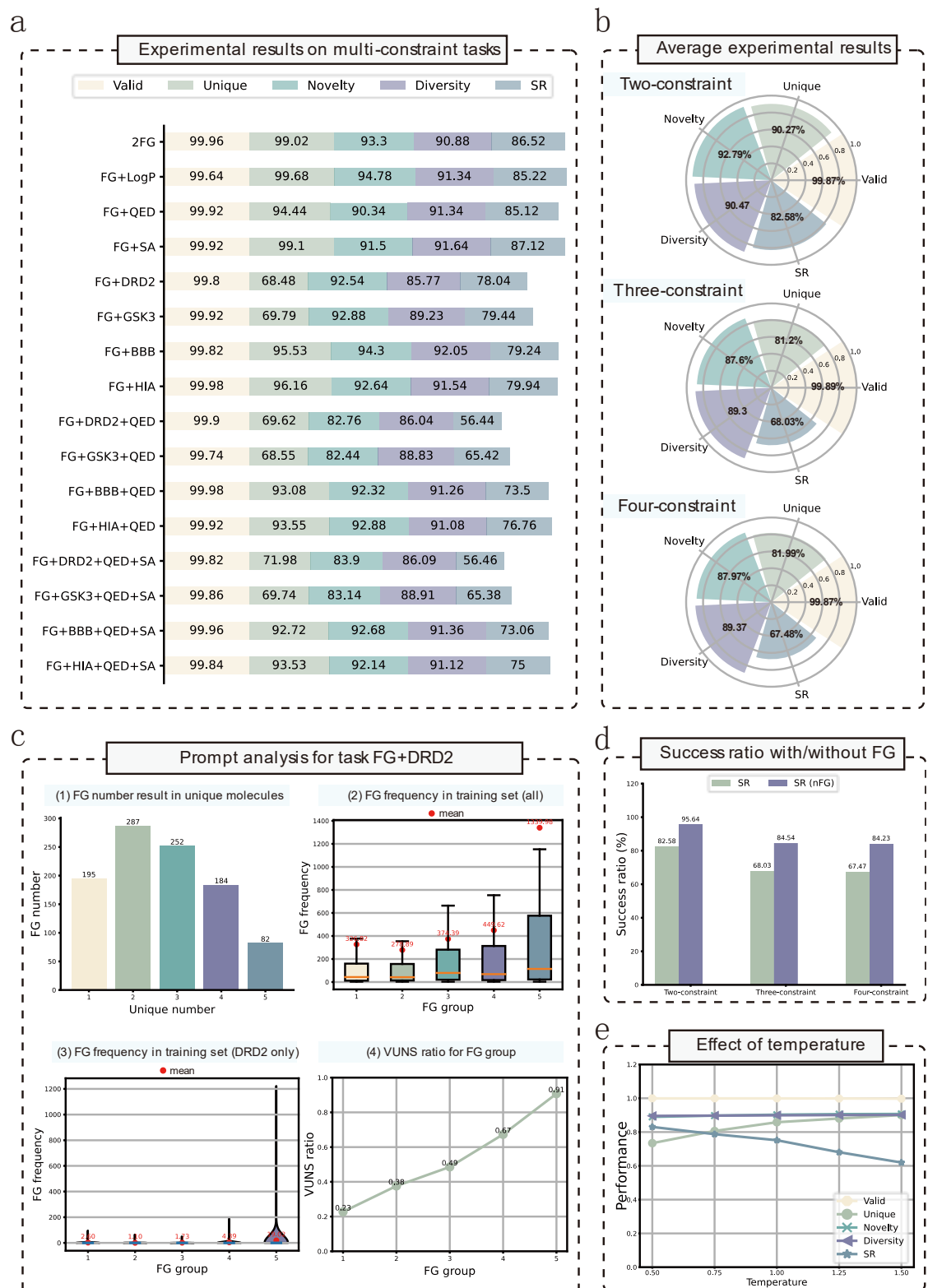

**Figure 2** (a) Experimental results for TSMMG across various tasks, encompassing 8 two constraint tasks, 4 three constraint tasks, and 4 four-constraint tasks. (b) Average experimental results on two-constraint, three-constraint, and four constraint tasks. (c) FG analysis for task FG+DRD2. (d) Shows the comparison of the success rate SR (nFG) without considering whether FG matches and the success rate considering whether FG matches under different constraint tasks. (e) Shows the impact of different temperatures on the model.

# Case Study of a Five-constraint molecular generation

Given the availability of corresponding predictors and a sufficiently large molecular library, it is theoretically feasible to construct training sets for any combination of desired molecular properties. This would enable the model to generate molecules that encompass a wide range of attributes. However, the challenge arises as the number of properties and their combinations increases, resulting in an exponential growth in the total number of possible property combinations. The exhaustive coverage of all these combinations becomes impractical. To address this challenge, we embarked on an investigation to determine if a model could effectively generate molecules when trained using individual properties but tested on arbitrary combinations. This research aimed to assess the model's adaptability to novel scenarios. To this end, we designed a task that entailed the generation of molecules exhibiting high drug-likeness, good synthetic accessibility, blood-brain barrier permeability, and the ability to bind to both the prostaglandin E2 receptor EP2 subtype [16] and prostaglandin E receptor EP4 [17]. The input prompt constructed for this task was: 'The molecule exhibits a high QED score, good synthetic accessibility. It can pass through the blood-brain barrier and binds to both Prostanoid EP2 and EP4 receptors.' During the training phase, each molecule was associated with only one property, meaning the model was exposed to molecules corresponding to each of the five properties within this prompt. However, the model had not encountered molecules that simultaneously met all five of these properties, and indeed, it had not seen molecules that met even two properties explicitly simultaneously.

This task presents a formidable challenge from multiple perspectives. From the model's input perspective, the model encounters significantly longer input text, a departure from its prior training data. In terms of molecular properties, the model must not only comprehend the mapping of individual properties to molecular spaces but also grasp the complex mapping of multiple properties from a single property mapping. Surprisingly, the model proves to be up to the task, successfully generating molecules that simultaneously satisfy all the condition constraints. As illustrated in Figure 3, we showcase four molecules that meet all the property requirements specified in the textual description. To validate their compatibility with the receptors of EP2 (PDB ID: 7CX2) and EP4 (PDB ID: 5YWY), we employed UCSF Chimera [18] for molecular docking preparation and UCSF Dock6 [19] to conduct molecular docking. Finally, we used PLIP [20] and PyMOL [21] for visualizing the docking results. The docking results reveal that these molecules effectively fit into the ligands and establish hydrogen bonds with different residues, demonstrating their potential for fulfilling the specified molecular properties.

This experimental outcome holds profound significance, as it demonstrates the model's robust capability to generate molecules that satisfy complex multi-constraints during zero-shot testing, even when initially trained with one-constraint data. This versatility underscores the model's adaptability and its potential to address intricate challenges in molecular generation.

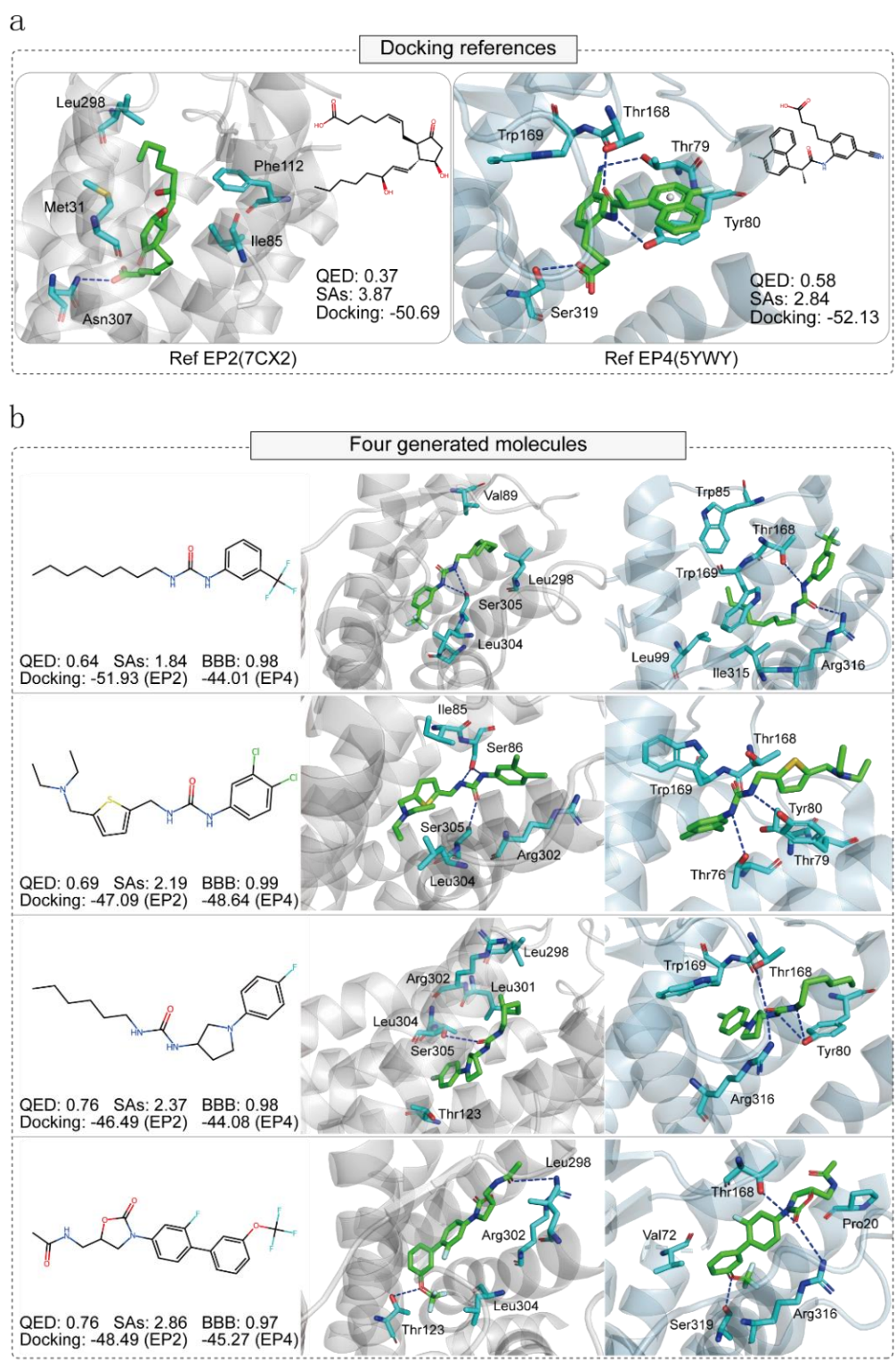

**Figure 3** (a) Docking reference for EP2 and EP4. (b) Molecules generated by TSMMG that can simultaneously bind to both EP2 and EP4 receptors. The input prompt is: "The molecule exhibits a high QED score, good synthetic accessibility. It can pass through the blood-brain barrier and binds to both Prostanoid EP2 and EP4 receptors." During training, TSMMG has seen molecules that can individually bind to both EP2 and EP4 receptors, but it has not explicitly received molecules that simultaneously satisfy all the constraints in this prompt. Nevertheless, it still successfully generates the desired molecules.

# Diversity of Input Text

Given that TSMMG is trained based on GPT-2[40], which has undergone extensive pre-training on natural language datasets, we have a reasonable basis to hypothesize that TSMMG can comprehend the similarities in natural language. Specifically, when provided with prompts that share the same semantics but exhibit subtle differences in their expressions, TSMMG is likely to generate accurate molecules. This hypothesis stems from the fact that GPT-2 has acquired the ability to capture various linguistic patterns and semantic relationships during its training process. Consequently, it may possess the capability to generalize and transfer its knowledge to related but slightly different prompts. In essence, TSMMG's potential to generate correct molecules may persist even with prompt variations due to its underlying understanding of linguistic similarities.

To test this hypothesis, we explored the use of diverse templates that encompass different language habits and variations. By making slight modifications to the original training templates, we aimed to assess TSMMG's ability to generate correct molecules when input prompts were slightly altered. For example, during the training phase, we utilized the template 'The molecule contains [FG], it can be absorbed by the human intestinal tract.' for the FG+HIA task. We introduced minor adjustments to create two new prompts: 'I want a molecule that contains [FG] and can be absorbed by the human intestinal.' and 'Give me a molecule which contains [FG] and can be absorbed by the human intestinal.' We conducted experiments using these diverse prompts across four different tasks, as presented in Table 2, and the results are summarized in Table 3.

**Table 2** Prompts we used in order to test the tolerance of TSMMG to diverse inputs, [FG] refers to any functional group.

| | | |
|---|---|---|
| DRD2/GSK3 | T0 | The molecule contains [FG]. It is active to DRD2. [D2D2/GSK3]. |
| | T1 | I want a molecule that contains [FG] and can bind to [D2D2/GSK3]. |
| | T2 | Give me a molecule which contains [FG] and can bind to [D2D2/GSK3]. |
| BBB | T0 | The molecule contains [FG]. It can pass through the blood-brain barrier. |
| | T1 | I want a molecule that contains [FG] and can pass through the blood-brain barrier. |
| | T2 | Give me a molecule which contains [FG] and can pass through the blood-brain barrier. |
| HIA | T0 | The molecule contains [FG]. It can be absorbed by human intestinal. |
| | T1 | I want a molecule that contains [FG] and can be absorbed by human intestinal. |
| | T2 | Give me a molecule which contains [FG] and can be absorbed by human intestinal. |

The experiments demonstrated that TSMMG consistently generated molecules that met the specified requirements to a large extent, even with modified prompts. As shown in the Table 2, the validity of the generated molecules can still reach over 99% after using prompts of different styles. For the FG+BBB and FG+HIA tasks, using the T1 and T2 templates both resulted in approximately a 9% decrease in SR compared to using the T0 template, while uniqueness, novelty, and diversity showed almost no significant changes. For the FG+DRD2 task, when using the T1 template, SR decreased by 33.36%, novelty decreased by 12.12%, while uniqueness increased by 3.36%; when using the T2 template, SR decreased by 30%, novelty decreased by 11.22%, while uniqueness increased by 1.58%. The FG+GSK3 task and the FG+DRD2 task show the same trend, that is, when using the T1 and T2 templates, SR and novelty shows a significant decrease and uniqueness shows a certain degree of increase, while other indicators show relatively small differences.

**Table 3** Experimental results with different template styles.

| template | task | valid | unique | novel | diversity | SR | SR (nFG) |
|---|---|---|---|---|---|---|---|
| T0 | FG+DRD2 | 99.80% | 68.48% | 92.54% | 85.77% | 78.04% | 93.18% |
| T0 | FG+GSK3 | 99.92% | 69.79% | 92.88% | 89.23% | 79.44% | 94.40% |
| T0 | FG+BBB | 99.82% | 95.53% | 94.30% | 92.05% | 79.24% | 96.14% |
| T0 | FG+HIA | 99.98% | 96.16% | 92.64% | 91.54% | 79.94% | 95.80% |
| T1 | FG+DRD2 | 99.70% | 71.84% | 80.42% | 86.26% | 44.68% | 82.12% |
| T1 | FG+GSK3 | 99.68% | 73.49% | 83.90% | 89.56% | 47.36% | 84.88% |
| T1 | FG+BBB | 99.68% | 94.30% | 95.48% | 92.28% | 70.64% | 96.84% |
| T1 | FG+HIA | 99.78% | 94.11% | 94.82% | 91.99% | 69.34% | 93.86% |
| T2 | FG+DRD2 | 99.80% | 70.06% | 81.32% | 86.18% | 48.04% | 83.52% |
| T2 | FG+GSK3 | 99.68% | 71.89% | 84.68% | 89.52% | 50.22% | 85.60% |
| T2 | FG+BBB | 99.74% | 95.67% | 95.40% | 92.17% | 70.72% | 96.52% |
| T2 | FG+HIA | 99.90% | 95.24% | 95.48% | 91.86% | 71.10% | 94.12% |

These results suggest that TSMMG exhibits a certain degree of tolerance to diverse prompts and can continue to generate molecules that meet the specified requirements, even when the prompts are modified. It is important to note that while TSMMG may demonstrate tolerance to prompt variations, the extent of its ability to generalize and generate accurate molecules may vary depending on the specific prompt and task.

# Discussion

## TSMMG as producer

The development of TSMMG can be viewed as a form of knowledge distillation [46], as depicted in Figure 4-a. Initially, diverse molecular properties are obtained using teacher models. These properties are then encapsulated into natural language descriptions and combined with molecular sequences to create text-molecule pairs. TSMMG is trained using these text-molecule pairs as training data, enabling it to acquire the knowledge inherent in the properties through natural language. By leveraging this process, TSMMG becomes proficient in generating molecules that exhibit the desired properties. Notably, TSMMG has the ability to generate novel molecules possessing specific properties based on the acquired knowledge. This offers a feedback loop to the teacher models, allowing them to refine and update their knowledge.

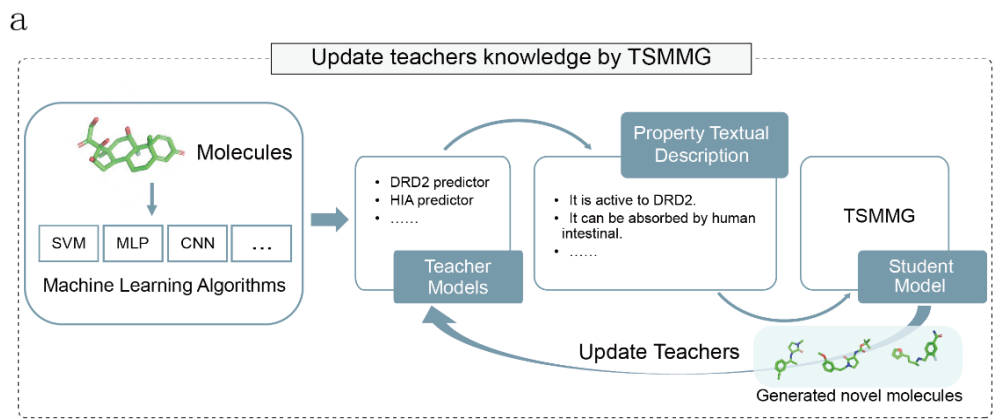

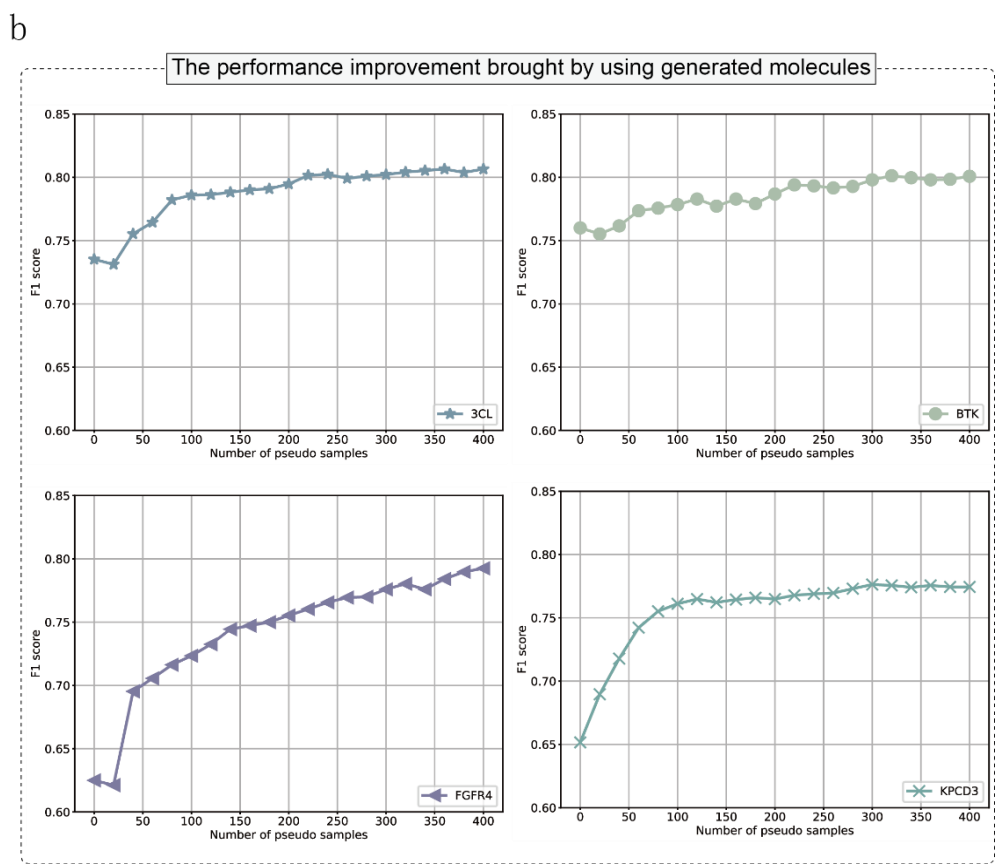

**Figure 4** TSMMG as a producer. Molecules generated by TSMMG can be used to improve the accuracy of the predictor. (a) We leverage a large number of property predictors, which can be regarded as teacher models, to obtain molecular properties, and then use these properties to construct textual descriptions to train TSMMG. The molecules generated by TSMMG can also be used to update the corresponding property predictors. This has two benefits: firstly, it allows us to verify whether TSMMG has effectively extracted the latent representation of the property, and secondly, it can improve the accuracy of the property predictors. (b) The experimental results on four property predictors are shown in the figure. The horizontal axis represents the number of generated molecules added to the training data, which we refer to as pseudo-samples. As can be observed, the accuracy of the property predictors increases and tends to converge as the number of pseudo-samples increases.

To illustrate this, we conducted further experiments involving Serine/threonine-protein kinase D3 (KPCD3), Bruton's tyrosine kinase (BTK), Fibroblast growth factor receptor 4 (FGFR4), and Papain-like protease 3CL. Initially, each target dataset is randomly divided into training and test sets. Subsequently, a random subset is partitioned from the training dataset to serve as the validation set, and an SVM predictor is trained using the training set and validated using the validation set. This process is repeated 100 times to select the best predictor. Finally, the chosen predictor is applied to the test set to obtain the F1 score. For comparison, different numbers of molecules generated by TSMMG that can bind to the corresponding target are randomly added to the training set to train new SVM predictors. This process is also repeated 100 times to yield consistent statistical data. These added molecules are referred to as pseudo samples. The experimental results presented in Figure 4-b demonstrate that the addition of pseudo samples significantly enhances the performance of the predictors. Notable improvements are observed in KPCD3, BTK, FGFR4, and 3CL by approximately 13%, 4%, 17%, and 7%, respectively. Furthermore, as the number of pseudo samples increases, the performance of each predictor tends to converge. These results indicate that TSMMG can discern the commonalities shared by molecules with specific properties and generate novel molecules that embody these commonalities. Moreover, TSMMG's unique ability to learn from teacher models and provide feedback for updating the knowledge of these models initiates a symbiotic relationship that promotes continuous improvement in their respective capabilities.

# Conclusions

In conclusion, the TSMMG model represents a significant leap forward in the field of multi-constraint molecular generation for drug discovery and related applications. TSMMG's innovative approach addresses several critical limitations in existing methodologies, unlocking the potential for more efficient and versatile drug discovery processes. Through a comprehensive set of experiments, TSMMG's efficacy is demonstrated across a range of tasks, from two-constraint to four-constraint molecular generation. The model consistently excels in generating molecules that adhere to predefined conditions, boasting impressive success rates across various property combinations. Moreover, TSMMG exhibits remarkable abilities when subjected to zero-shot testing, generating molecules that fulfill complex multi-constraint requirements, even when such combinations were not present in the training data. This feature demonstrates TSMMG's adaptability and its potential to revolutionize molecular design.

TSMMG offers a fresh perspective and promising capabilities, paving the way for more streamlined, efficient, and adaptable approaches. Its potential to expand into other domains, such as materials science, signifies the wide-reaching impact of this innovative model. TSMMG stands as a testament to the synergy between natural language understanding and molecular generation, opening new doors for researchers in the quest for novel compounds and therapies.

# Method

## Problem Setting

Natural language serves as a user-friendly means for human-machine interaction, making it an ideal solution for generating molecules from natural language descriptions. Recent successes in the development of large language models (LLMs) [22][23] inspire the vision that we may achieve the generation of molecules from diverse molecular spaces by simply modifying the input prompt. This approach offers a promising solution to address the challenge of generality in molecular generation. Despite both natural language and SMILES being sequence data formats, SMILES can be viewed as a specialized molecular language that can be challenging for humans to interpret. From this perspective, generating molecular sequences from natural language descriptions can be regarded as a translation task, an area where LLMs excel.

Given a natural language sequence, $W = \{w_1, w_2, \ldots, w_n\}$, the objective is to generate a corresponding molecule represented by a SMILES sequence, $S = \{s_1, s_2, \ldots, s_m\}$, which can be formulated as conditional probability:

$$P(S|W)$$

In order to ensure the quality of the generated molecules, it is imperative to adhere to the following prerequisites: (1) **Validity**. The generated SMILES representation, $S$, should strictly adhere to the syntax rules of the SMILES format, guaranteeing that it forms a valid and well-structured molecule. (2) **Relevance**: The molecule represented by $S$ should accurately reflect the physical and chemical properties described by the natural language sequence $W$. This entails that if there exists a subsequence $W_{i,j}$ in $W$ that specifies a particular property, there should be a corresponding subsequence $S_{k,l}$ in $S$ that satisfies the desired property. (3) **Diversity**: While satisfying the validity and relevance criteria, the generated $S$ should exhibit diversity. In other words, the generated molecules should not be identical or overly similar, providing a range of molecular structures that fulfill the given properties. (4) **Novelty**: The model should possess the ability to generate $S$ that are not present in the training set. This capability ensures that the generated molecules introduce new and previously unseen chemical structures, thereby expanding the exploration space beyond the confines of the training data.

The quandary of translating natural language into molecular language, albeit bearing resemblances to conventional machine translation, poses distinctive challenges. In this context, three fundamental patterns of correspondence between natural language and molecular sequences can be discerned: (1) **One-to-One Mapping**: In this pattern, a specific text description corresponds to a single, specific molecular sequence. Models like MolT5 [24], MolXPT [25], and MoleculeSTM [26] have tackled this problem as a query task, aiming to establish a direct mapping relationship between text and molecular sequences. However, this approach may not be ideal for generating novel molecules with diverse properties, as it relies on a fixed ground truth and does not explore beyond the known data.

(2) **One-to-Many Mapping**: Here, a text description can correspond to multiple different molecular sequences. This pattern allows the model to learn the feature distribution of the target space, enabling sampling from the distribution to generate new molecules. Models like those proposed by Kotsias et al. [27] and Wang et al. [10] leverage this pattern effectively by training on specific datasets containing molecules with shared properties which implicitly embracing the One-to-Many mapping pattern. (3) **Many-to-One Mapping**: In this pattern, a specific molecular sequence can be described in various ways. By understanding the inherent relationship between different attributes, it is possible to discover new properties of a molecule. This pattern offers opportunities for exploring diverse attributes of molecules beyond their known properties.

In order to develop a universal molecular generative model capable of generating molecules with various desired properties without the need for retraining, it is essential to accumulate a substantial amount of data that explicitly adheres to the One-to-Many and Many-to-One mapping patterns. The primary challenge lies in acquiring a sufficient number of text-molecule pairs in a rapid, convenient, and cost-effective manner.

## Data Generation Framework

Several studies have explored the integration of natural language and molecular language. MolT5 [24], aimed to achieve bidirectional translation between natural language and molecular language. The model underwent initial pre-training on an extensive collection of unpaired natural language corpora and molecular sequences, followed by fine-tuning on the text-molecule paired dataset ChEBI-20. However, ChEBI-20 presents two notable limitations. **Firstly**, it contains a relatively small set of 33,010 text-molecule pairs, making it challenging to establish the correspondence between natural language and molecular language. **Secondly**, the text descriptions in this dataset, sourced from the comment field in ChEBI [28], often contain information unrelated to molecular properties. Additionally, these descriptions exhibit a strong one-to-one relationship with the molecules, posing challenges for the model to explore the specific molecular space associated with desired properties. MolXPT [25] proposed a method that involves incorporating molecular sequences within the input text for Large Language Models (LLMs). CLAMP [29] introduced a fusion approach, combining a molecule encoder and a text encoder for property prediction tasks. Christofidellis et al. [30] presented a unified model capable of handling various text-to-text, text-to-molecule, molecule-to-text, and molecule-to-molecule tasks. MolReGPT [31] implemented tasks such as molecule captioning and text-based molecule generation by assigning ChatGPT a role as a biochemist, facilitating in-context learning.

However, a common limitation in all of the above-mentioned studies is their reliance on the ChEBI dataset, which constrains their performance due to data scarcity and quality issues. As of now, limited research efforts have been directed at addressing these issues in natural language-based molecular generation. Therefore, we propose a knowledge distillation-based approach to construct an extensive and high-quality dataset of natural language-molecule pairs.

Figure 1-a provides an overview of the framework employed for the creation of our dataset. The underlying concept revolves around the utilization of advanced molecular parsing tools and models

to extract knowledge related to molecules. Subsequently, this acquired knowledge is transformed into natural language text, resulting in paired data comprising molecules and their corresponding textual descriptions. Within this framework, the tools and models responsible for extracting molecular knowledge are collectively referred to as 'teachers', while TSMMG assumes the role of the 'student'. TSMMG undertakes the task of learning various properties associated with molecules from these 'teachers'. It also comprehends the mapping relationship between these properties and the molecular structures themselves. This knowledge empowers TSMMG to generate new molecules based on specified properties using natural language descriptions.

Within this framework, multiple 'teachers' are employed, each with distinct capabilities related to molecular properties and structures. These teachers encompass a range of tools and models, including:

**Physicochemical Property Teacher:** RDKit, a tool capable of parsing molecules to extract physicochemical properties such as molecular weight (MW), the number of aromatic rings (AROM), LogP, SA, QED, the number of hydrogen bond acceptors (HBA), the number of hydrogen bond donors (HBD), and topological molecular polar surface area (PSA).

**ADMET Property Prediction Models:** admetSAR [32], based on Support Vector Machine (SVM), predicts ADMET properties, such as blood-brain barrier permeability and absorptivity.

**Affinity Prediction Models:** Olivecrona et al.'s SVM-based models [33] and Jin et al.'s models [34] can predict the binding probabilities of molecules to specific targets, including DRD2, GSK3, and JNK3. Newer models such as MolTrans [35], DrugBAN [36], and TransformerCPI [37] are designed to predict the affinity of small molecules to receptor proteins and more.

**Structural Information Extraction:** In addition to these property-related teachers, the IUPAC name of a molecule, which bears structural information, is considered. The IUPAC name exhibits a grammar resembling natural language and provides standardized descriptions of molecules. By breaking down IUPAC names, it is possible to extract structural components of a molecule. For instance, deconstructing the molecule '(2-methyl-5-methylsulfonylphenyl)methanamine' yields the functional groups 'methyl,' 'methylsulfonylphenyl,' and 'methanamine.' Therefore, an IUPAC parser is proposed, along with a set of rules for dissecting IUPAC names, serving as an additional 'teacher' for extracting the internal structure of molecules.

Through these 'teachers,' we acquire extensive knowledge about molecules, including their structural information, physicochemical properties, and binding affinities to specific receptors. This information is then transcribed into natural language descriptions and combined with the corresponding molecules to create text-molecule pairs. For an example as shown in Figure 1-b, let's consider a molecule represented as 'CCN1CCCC1CNC(=O)c1c(OC)ccc(Cl)c1O.' We can break down its IUPAC name to extract the functional group 'methoxybenzamide.' By utilizing RDKit, we determine its LogP, QED, and SAs. We further predict its affinity with DRD2 through a classifier proposed by Olivecrona et al. [33], and evaluate its likelihood of passing through the blood-brain barrier using admetSAR. These various properties are then associated with the molecule using natural language templates.

The data generation method offers several notable advantages: (1) With numerous publicly accessible molecular databases like PubChem [38] and ZINC [39], our approach allows for the rapid acquisition of a large number of text-molecule pairs. This effectively overcomes the data limitations often encountered in natural language-based molecular generation models; (2) The text molecule pairs generated through this method exhibit a high degree of relevance. Each segment of text contains certain properties of the molecules, enabling the model to learn the mapping relationship between text descriptions and molecular properties more effectively; (3) There is a wealth of advanced tools and models available for molecular structure analysis and property prediction. Our framework simplifies the process of transferring knowledge from these advanced tools and models into a student model in natural language form. This empowers the student model to generate molecules that incorporate this knowledge. (4) The method is highly scalable, allowing for the seamless transfer of knowledge for various molecular properties. It can be applied to an array of properties, making it versatile for different research needs. (5) Our method supports continuous knowledge updates. This means that the student model can benefit from the latest and more robust models, ensuring that it remains up-to-date and well-informed.

## Training Model

We began by collecting 2 million molecules from PubChem. Subsequently, we harnessed the tools and models mentioned earlier to extract comprehensive knowledge regarding these molecules. This knowledge was then translated into natural language text using predefined templates and combined with the corresponding SMILES representations. To maximize the model's capabilities, we thoughtfully organized the data to encompass both One-to-Many and Many-to-One patterns. This approach ensures that the model learns the underlying distribution of specific inputs, promoting adaptability and preventing the mere memorization of fixed responses.

For instance, let's take molecule $M$, which possesses ten pieces of extracted knowledge. If we were to compile all ten pieces into a single text, denoted as $T$, the resulting molecule space associated with $T$ would likely be highly restricted, potentially corresponding to just one specific molecule, let's say, molecule $A$. This would essentially create a One-to-One data pattern. To overcome this limitation, we adopt a strategy where, for each molecule, we selectively choose a subset of its knowledge to compose the text. The goal here is to craft this text in a way that it corresponds to as many molecules as possible. This method empowers the model to gain insights into the broader distribution of molecules linked to the provided text, rather than locking it into a specific, isolated instance.

Then the training of TSMMG involves two key steps: pre-training on a large natural language corpus and fine-tuning on text-molecule paired data that we have constructed. In the first stage, TSMMG undergoes pre-training on a large natural language corpus. This enables TSMMG to learn and understand natural language by capturing the statistical patterns and linguistic structures present in the data. The pre-training stage helps TSMMG acquire a general understanding of language and forms the initial foundation for subsequent training stages. The second stage involves a fine-tuning on the text-molecule paired data that contains descriptions of various properties as shown in Tabel 3. This fine-tuning stage focuses on teaching TSMMG the mapping between text descriptions and molecular sequences as well as the syntax of SMILES. By fine-tuning TSMMG on this specific dataset, it becomes proficient in generating molecules based on specific text-described-property such as functional groups, LogP, physicochemical properties, drug-like properties, and affinity scores to certain targets.

The architecture of TSMMG is the same as GPT [40]. TSMMG follows the settings of GPT2$_{small}$, which consists of 12 layers and has a total of 117 million parameters. We downloaded the weights of GPT2$_{small}$ from Huggingface model repository [47] to initialize TSMMG. This helps with cost and computational considerations by leveraging pre-trained weights for an efficient starting point. And since the weights are trained by a large number of language corpus, we can directly fine-tune the model using the text-molecule paired data we construct. We fine-tune TSMMG on 8 A100 40G GPUs for around 6 days. We use the subsequent hyperparameters: a batch size of 32, a learning rate set to 5e-4, a warmup steps of 100. We use AdamW [42] as the optimizer.

## Metrics

To evaluate the performance of the TSMMG model, we employed four common metrics in molecular generation: **Validity**, **Uniqueness**, **Novelty** and **Diversity**. Each of these metrics was essential for a comprehensive evaluation: **Validity** assesses whether the generated molecules conform to the syntax rules of SMILES. We utilized RDKit [43] to parse the generated molecules, considering them valid if the parsing process was successful. **Uniqueness** measures the proportion of non-repetitive molecules among the generated set. It ensures that the model produces diverse molecules. **Novelty** signifies whether the generated molecules are previously unseen in the training dataset, preventing the model from regenerating known molecules. **Diversity** describes the structural differences between generated molecules, it is calculated as:

$$Diversity = 1 - \frac{2}{n(n-1)}\sum_{X,Y} sim(X,Y).$$

Where $sim(X, Y)$ is calculated based on the Tanimoto distance with respect to the Morgan fingerprints of generated molecules $X$ and $Y$.

In addition to these standard metrics, we introduced the concept of success rate (**SR**) to measure whether the generated molecules meet predefined conditions. We establish different criteria to define the success of generated molecules based on the specific task. These criteria are outlined as follows:

**FG**: We leveraged IUPAC nomenclature to identify functional groups within the molecules. By parsing IUPAC names and matching them to generated SMILES-encoded molecules, we checked if the generated molecules contained the specified functional groups.

**LogP**: Using RDKit, we calculated the LogP values of the generated molecules and compared them to predefined values. The generation was considered successful if the LogP value fell within a margin of 1 from the specified value.

**QED and SAs**: For these tasks, we adopted criteria similar to prior work [10], considering QED as high if its value exceeded 0.6 and SAs as good if the score was less than 4.

**DRD2 and GSK3**: We employed the models proposed by Jin et al. [34] to predict the affinity scores of the generated molecules. A molecule was considered successful if its corresponding affinity score exceeded 0.5 for either target.

**BBB and HIA**: We used models developed by Cheng et al. [32] to predict scores, determining if a molecule could pass through the blood-brain barrier (BBB) if its BBB score exceeded 0.5 or if it could be absorbed by the human small intestine (HIA) if its HIA score was above 0.5.

Moreover, for each multi-constraint task, we only considered molecules successful if they met all constraints contained in this task simultaneously. We generated 5000 molecules to evaluate the model's performance for each multi-constraint task. Note that we uniformly express all metric results in percentage format. While converting **Diversity** to a percentage may lack intrinsic meaning, for the sake of ease of comparison with other metrics, we multiply it by 100. However, we refrain from appending the '%' symbol to distinguish it from other metrics.

## Translating SMILES to IUPAC

There are several open works that provided their solutions for translating SMILES to IUPAC name, such as STOUT [44] and IUPAC2Struct [45], but the interfaces they released are not for high throughput experiments. Considering experimental efficiency, we trained our own SMILES2IUPAC model based on GPT2$_{small}$. We formulate this problem also as conditional probability $P(I|S)$ where generating a corresponding IUPAC name $I = \{i_1, i_2, ..., i_n\}$ by a given SMILES sequence $S = \{s_1, s_2, ..., s_m\}$. We collect 2 million SMILES-IUPAC paired data from Pubchem to train this model. The model size and settings of SMILES2IUPAC are the same as TSMMG. For evaluating the trained SMILES2IUPAC model, we pass 1000 unseen molecules to it and generate 1000 corresponding predicted IUPAC names. We then break down the predicted IUPAC names to identify the functional groups, and check if all these functional groups exist in the corresponding ground truth IUPAC names. The experimental results show an accuracy rate of 94%.

# Declarations

## Ethics approval and consent to participate

This study involves computational experiments that are non-invasive and do not directly intervene with any human or animal subjects. Therefore, ethical approval from an institutional review board is not required. The consent of participants and the protection of personal information are not applicable, as the experimental data are sourced from publicly available datasets.

## Consent for publication

All authors have provided their consent for publication of this study. There are no identifiable individuals or personal data included in this manuscript, ensuring compliance with publication ethics.

## Availability of data and materials

Source data and codes are provided with this paper. The datasets used for training and evaluating TSMMG are available at: https://github.com/HHW-zhou/TSMMG.

## Competing interests

The other authors have declared no competing interests.

## Funding

This work was partly supported by the National Natural Science Foundation of China (62425204, 62122025, U22A2037, 62450002, 62432011).

## Author contributions

L.W., YS.L., and X.Z. conceived the study of TSMMG and the experimental assays. P.Z. developed TSMMG. P.Z., J.W., C.L., and Z.W. performed all experiments. X.Z., YS.L., L.W., and P.Z. drafted the manuscript and charts. Y.L., C.L., S.S, J.L., L.W., X.C., H.L., and W.L. critically revised the manuscript. All authors critically revised and gave final approval of the manuscript.

## Acknowledgments

We would like to thank the National Natural Science Foundation of China for their support of this work (No. 62425204, 62122025, U22A2037, 62450002, 62432011). We also appreciate the contributions of all authors involved in this study, whose collaboration made this research possible.